\documentclass[10pt,twocolumn,letterpaper]{article}

\usepackage{cvpr}              

\usepackage[dvipsnames]{xcolor}

\definecolor{cvprblue}{rgb}{0.21,0.49,0.74}
\usepackage[pagebackref,breaklinks,colorlinks,citecolor=cvprblue]{hyperref}

\usepackage{amsmath}
\usepackage{amssymb}
\usepackage{booktabs}
\usepackage{multirow}
\usepackage{tabulary}
\usepackage{makecell}
\usepackage{float}                  
\usepackage{overpic} 
\usepackage{indentfirst}

\def\paperID{9237} 
\def\confName{CVPR}
\def\confYear{2024}

\title{KD-DETR: Knowledge Distillation for Detection Transformer with Consistent Distillation Points Sampling}

\author{Yu Wang, Xin Li, Shengzhao Weng, Gang Zhang, Haixiao Yue, \\Haocheng Feng, Junyu Han, Errui Ding\\
Baidu VIS\\
{\tt\small \{wangyu106, lixin68, wengshengzhao, zhanggang03,}\\
{\tt\small yuehaoxiao, fenghaocheng, hanjunyu, dingerrui\}@baidu.com}
}

\begin{document}
\maketitle
\begin{abstract}
DETR is a novel end-to-end transformer architecture object detector, which significantly outperforms classic detectors when scaling up. In this paper, we focus on the compression of DETR with knowledge distillation. While knowledge distillation has been well-studied in classic detectors, there is a lack of researches on how to make it work effectively on DETR. We first provide experimental and theoretical analysis to point out that the main challenge in DETR distillation is the lack of consistent distillation points. Distillation points refer to the corresponding inputs of the predictions for student to mimic, which have different formulations in CNN detector and DETR, and reliable distillation requires sufficient distillation points which are consistent between teacher and student. 

Based on this observation, we propose the first general knowledge distillation paradigm for DETR (KD-DETR) with consistent distillation points sampling, for both homogeneous and heterogeneous distillation. Specifically, we decouple detection and distillation tasks by introducing a set of specialized object queries to construct distillation points for DETR. We further propose a general-to-specific distillation points sampling strategy to explore the extensibility of KD-DETR. Extensive experiments validate the effectiveness and generalization of KD-DETR. For both single-scale DAB-DETR and multis-scale Deformable DETR and DINO, KD-DETR boost the performance of student model with improvements of $2.6\%-5.2\%$. We further extend KD-DETR to heterogeneous distillation, and achieves $2.1\%$ improvement by distilling the knowledge from DINO to Faster R-CNN with ResNet-50, which is comparable with homogeneous distillation methods.The code is available at \url{https://github.com/wennyuhey/KD-DETR}
\vspace{-10pt}
\end{abstract}
    
\vspace{-12pt}
\section{Introduction}
\label{sec:intro}
 \begin{figure}[t]
	\centering
	\subfloat[Distillation Points in Classic Detector]{\includegraphics[width=0.7\columnwidth]{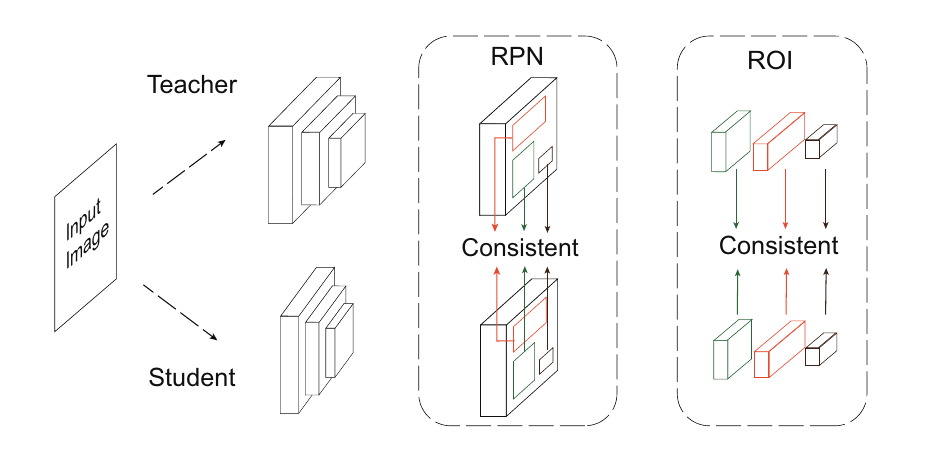}\label{classic}}\\
	\subfloat[Distillation Points in DETR]{\includegraphics[width=0.7\columnwidth]{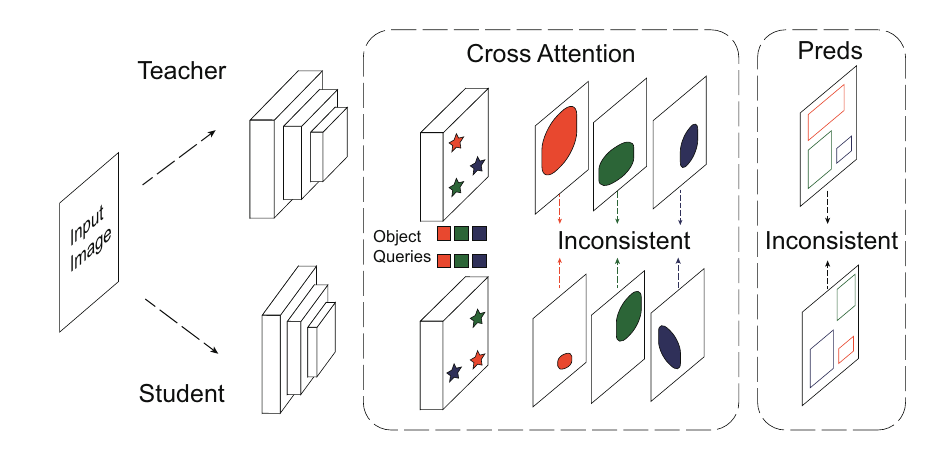}\label{detr}}\\
	\subfloat[Distillation Points in KD-DETR]{\includegraphics[width=0.7\columnwidth]{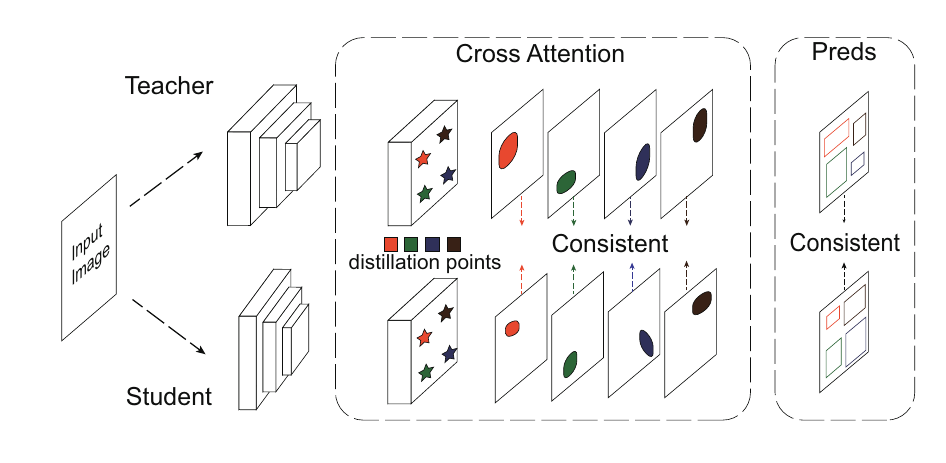}\label{kddetr}}
	\caption{\textbf{Schematic Illustrations of Distillation Points in Different Architecture}: (a) Two-Stage Detector: both positive and negetive proposals in RPN and RoI are consistent distillation points with strict one-to-one correspondence between teacher and student model; (b) DETR: Object queries lacks spatial or semantic relationship between teacher and student model, resulting in inconsistent distillation points; (c)In KD-DETR, a set of special object queries is introduced to construct consistent distillation points for DETR distillation}
	\vspace{-12pt}
\end{figure}

In recent years, \cite{carion2020end} propose the novel end-to-end detector Detection Transformer (DETR) which eliminates the need for hand-crafted anchors and non-maximum suppression (NMS). \cite{zhu2020deformable}\cite{li2022dn}\cite{liu2021dab}\cite{zhang2022dino}further make remarkable stride towards the scalability and potential of DETR, significantly outperforming classical detectors.

Different from classic detectors, DETR interprets object detection as an end-to-end set prediction problem with bipartite matching. A set of learnable object queries are introduced, each responsible for a certain instance. The object queries interact with features extracted from the encoder to make final predictions of box locations and categories. Despite the impressive performance, the growing model scale prevents DETR from being deployed to real-world applications with urgent computation budget requirement.

To address this problem, current works have made efforts in designing efficient DETR architectures, reducing the encoder tokens utilized in cross-attention module to decrease the computation cost\cite{roh2021sparse}, and leveraging dense prior from RPN to downsize the decoder layers \cite{yao2021efficient}. In this work, we concentrate on compressing the large-scale DETR model by knowledge distillation\cite{hinton2015distilling} approaches, which is a promising technique for model compression and accuracy boosting. Knowledge distillation can transfer the knowledge learned from large and cumbersome DETR models to small and efficient ones by forcing the student to mimic the predictions from teacher, whether logits or internal activations. However, modern knowledge distillation methods are designed under CNN-based detectors, and researches on expanding it to general DETR compression are limited. 

 We start from experiments of applying classic logit-based distillation\cite{chen2017learning} to DETR to investigate the key point in DETR distillation. With these experiments, we observe that the critical challenge lies in the different formulations of DETR and classic detectors. Compared with classic detectors, the set-prediction formulation of DETR naturally contains fewer consistent distillation points. We use distillation points to denote the corresponding input of the prediction for mimicking in knowledge distillation, and the sufficiency and consistency of distillation points form the foundation of knowledge distillation. Specifically, abundant distillation points which are kept consistent between teacher and student models are essential for effective distillation. As shown in Figure \ref{classic}, classic detectors make predictions for a set of region proposals generated from the sliding window locations on the images with handcraft scales. This pattern naturally ensures a strict spatial correspondence between the large number of proposals made by teacher and student models, even for those negative ones with low confidence, thus providing a sufficient number of consistent distillation points for mimicking. In DETR, as is shown in Figure \ref{detr}, the distillation points actually consist of both image and object queries. However, the object queries from teacher and student models are egocentric and even differ in number, thus lacking definite correspondence, especially those redundant negative queries in bipartite matching. As the distillation points in DETR are inconsistent and insufficient, the predictions acquired from teacher are not reliable or informative for student to mimic.

The observation above raises the issue: how to obtain sufficient and consistent distillation points for DETR distillation. Previous work\cite{chang2023detrdistill} explicitly alleviate this issue by utilizing the bipartite matching between the object queries from teacher and student. However, the bipartite matching is not stable\cite{li2022dn}, and the matched object queries are just similar but not consistent, lacking sufficiency and extensibility. To directly address this issue, we propose a general knowledge distillation paradigm for DETR (KD-DETR) with consistent distillation points sampling. In KD-DETR, as illustrated in Figure \ref{kddetr}, we decouple detection and distillation task by introducing a set of specialized object queries to construct distillation points. The distillation points are unlearnable and shared between teacher and student models, probing the “dark knowledge” in teacher model. In this way, consistent distillation points with customized quantities become available. With the paradigm of KD-DETR, we propose a general-to-specific distillation points sampling strategy to probe comprehensive knowledge in teacher model. We further propose a coordination-based distillation points sampling strategy to extend KD-DETR to heterogeneous distillation between DETR and CNN detector.


To the best of our knowledge, this is the first work to propose a general knowledge distillation paradigm for DETR for both homogeneous and heterogeneous distillation. In this paper, we first provide a thorough analysis of the key points in DETR distillation. Based on the analysis, we design a novel KD-DETR which significantly improves the performance of DETR distillation. KD-DETR has both flexibility to different DETR architectures and potential for scalability, even for  heterogeneous distillation between DETR and CNN detectors. We conduct extensive experiments on the MS COCO2017\cite{lin2014microsoft} dataset on both homogeneous and heterogeneous distillation, and significantly boosts the performance of student models. DAB-DETR with ResNet-18 and ResNet-50 backbone achieves 41.4$\%$, 45.7$\%$ mAP, respectively, which are 5.2$\%$, 3.6$\%$ higher than the baseline. Deformable DETR with ResNet-18 and ResNet-50 reaches 43.7$\%$ and 48.3$\%$ mAP, 3.6$\%$ and 3.8$\%$ higher than the baseline, and outperform DETRDistill\cite{chang2023detrdistill} with 1.7$\%$ improvement. DINO with ResNet-18 and ResNet-50 also gains 4.4$\%$ and 2.6$\%$ improvement than baseline. We further extend KD-DETR to heterogeneous distillation, and achieves 2.1\% improvement by distilling the knowledge from DINO to Faster R-CNN with ResNet-50, which is comparable with homogeneous distillation methods.

\section{Related Work}
\subsection{Classic Object Detection}

Classic detectors with CNN view object detection as a verification task with a sliding window on the image to generate anchors. The mainstream detectors can be divided into one-stage detectors\cite{tian2019fcos}\cite{li2020generalized} and two-stage detectors\cite{ren2015faster}\cite{cai2019cascade}. One-stage detectors, such as Retinanet\cite{lin2017focal}, YOLO\cite{redmon2016you}, and FCOS\cite{tian2019fcos}, directly predict the category and regression of anchors on each pixel of the feature maps. While two-stage detectors such as Faster-RCNN\cite{ren2015faster} and its variants\cite{pang2019libra}\cite{yan2019meta}\cite{zhang2019cad} introduce a Region Proposal Networks (RPN) to generate proposals, and a ROIPool or ROIAlign\cite{wang2019distilling} to extract features of each region proposal for further classification and regression refinement. Both one-stage and two-stage detectors require post-processing, such as NMS, to remove duplicate predictions.

\subsection{Detection Transformer}

\cite{carion2020end} first propose an end-to-end transformer-based detector without any post-processing. Different from classic object detection, DETR interprets object detection as a set-prediction problem with bipartite matching.  Lots of follow-up focus on the slow convergence of DETR\cite{dai2021dynamic}\cite{sun2021rethinking}\cite{gao2021fast}\cite{zhang2022accelerating}. Deformable DETR\cite{zhu2020deformable} introduces a deformable attention module by generating reference points for query elements, each of which only concentrates on a small number of locations on the whole feature map. An alternative way is to add more prior information to the object queries in the decoder. Conditional-DETR\cite{meng2021conditional} decouples the context and position features in object queries and generates position features by spatial location. DAB-DETR\cite{liu2021dab} further introduces the width and height information to the positional features. Anchor DETR\cite{wang2022anchor} also encodes the anchor points as object queries with multiple patterns, and further designs a row-column decouple attention to reduce memory cost. The recent work of DINO\cite{zhang2022dino} draws the existing novel techniques, and further exerts the potential of DETR by enlarging the scale of model and datasets.

Besides, another problem in DETR is the scale and computation cost of the model. Current works solving this problem by designing more efficient DETR architecture. Sparse DETR\cite{roh2021sparse} reduces the computation cost by sparsifying encoder tokens. Efficient DETR\cite{yao2021efficient} otherwise introduce RPN to generate object queries and eliminate the cascading decoder layers in DETR. PnP DETR\cite{wang2021pnp} shorten the length of sampled feature with a poll and pool sampling module.

\subsection{Knowledge Distillation}

Knowledge distillation is a widely-used method for model compression and accuracy boosting by transferring the knowledge in a large cumbersome teacher model to a small student. \cite{hinton2015distilling} first propose the concept of knowledge distillation, where the student mimics the soft predictions from teacher. Knowledge distillation has been utilized in various fields\cite{tian2019contrastive}\cite{yang2022focal}\cite{yang2022cross}. According to the objective of mimicking, knowledge distillation can be divided into three categories: response-based\cite{zhao2022decoupled}, feature-based\cite{heo2019comprehensive}\cite{yang2022masked} and relation-based\cite{yim2017gift}\cite{zagoruyko2016paying}, which distill with logits, intermediate activations and the relation of features in different layers respectively.

Several works focus on applying knowledge distillation to object detection\cite{guo2021distilling}\cite{chen2017learning}\cite{kang2021instance}. \cite{chen2017learning} successfully distills the features on the neck, the classification head, and the regression head, while \cite{li2017mimicking} chooses to distill the logits and features from the RPN head. To overcome the imbalance of foreground and background, \cite{wang2019distilling} introduces fine-grained mask to focus on the regions close to ground-truth bounding boxes, \cite{dai2021general} pays more attention to the regions where teacher and student are divided in predictions. 

However, the modern knowledge distillation methods for object detection are built upon the architecture of CNN-based detectors, and are not suitable for DETR due to the completely different transformer architecture. \cite{chang2023detrdistill} directly introduces response-based and feature-based distillation to DETR with Hungarian-matching. Different from the previous work, we analyze the limitation of the set-prediction formulation in knowledge distillation, and propose a general paradigm for both homogeneous and heterogeneous DETR distillation.

\section{A Closer Look at DETR Distillation}

In this section, we first revisit the DETR architecture briefly. Then we conduct a series of classic knowledge distillation experiments on DETR to reveal that the core of DETR distillation is to obtain sufficient and consistent distillation points.

\subsection{Revisiting DETR}
DETR is built upon the encoder-decoder architecture of transformer. The encoder takes pixels of the feature map from backbone as input for multi-head self-attention to extract context features $X\in\mathbb{R}^{HW\times D}$, where $HW$ denotes the resolution of the feature map, and $D$ denotes the feature dimension. The decoder takes the features from encoder and a set of learnable object queries $\mathbf{Q}=\{q_i\in\mathbb{R}^D|i=1,...,N\}$ as input, where N denotes the number of queries. Each object query is an abstract feature describing a certain instance, and will probe and pool the features from encoder through cross-attention to make predictions of category $\mathbf{C}=\{\mathbf{c}_i\in\mathbb{R}^K|i=1, ..., N\}$ and location $\mathbf{B}=\{\mathbf{b}_i=(bx_i, by_i, bw_i, bh_i)|i=1, ..., N\}$, where K denotes the number of categories. Finally, the Hungarian algorithm is used to find a bipartite matching between ground truth and predictions of object queries. 
\begin{table}[t]
\begin{center}
\begin{tabular}{c|ccc}
\toprule[2pt]
Strategy & AP & $AP_{50}$ & $AP_{75}$  \\
\hline
Baseline & 36.2 &56.1 &37.9 \\
Inconsistent  &35.1 & 55.2& 36.7\\
Similar Foreground  &37.2 & 57.4 & 39.9  \\
Similar General & 37.4 & 58.0 &  40.6\\
\bottomrule[2pt]
\end{tabular}
\end{center}
\vspace{-5pt}
\caption{Distillation with Different Distillation Points. Inconsistent distillation points are harmful for distillation. }
\vspace{-15pt}
\label{inconsistent}
\end{table}

\subsection{Consistent Distillation Points}
The core idea of knowledge distillation is forcing the student to mimic the prediction of teacher, which can be interpreted as matching the mapping function of student and teacher with a set of distillation points. Distillation points refer to the corresponding input $\mathbf{x}$ of the predictions, as $\mathbf{y}=f(\mathbf{x})$, where $f$ represents the model. In this view, the distillation points should kept sufficient in quantity and consistent between teacher and student models for effective and reliable matching. However, comparing the formulation of CNN detector and DETR, a critical challenge of DETR distillation lies in the lacking of consistent distillation points.

Classic detector degrades object detection to a verification problem which combines classification and regression, and introduces a set of anchors to specify the region for verification. In this way, the formulation of distillation points consists of the image and the location and scale of anchor $\mathbf{x} = (\mathbf{I}, \mathbf{anchor})$. As the anchors are generated through the sliding window strategy with handcrafted shapes, the locations and scales of anchors are implicitly embedded in the model architecture as prior information. Since student and teacher models share the same or similar architectures, a large number of object proposals generated by teacher and student models naturally have strict spatial correspondence, even for those background regions with low confidence. With the spatial correspondence which can be viewed as an inductive bias of CNN, classic detector naturally guarantees a sufficient number of consistent distillation points. 

In contrast, the DETR formulates object detection as a set-prediction problem. The distillation points, therefore, become the combination of the image and the object queries $\mathbf{x} = (\mathbf{I}, \mathbf{q})$. However, the object queries in different models are egocentric, as they are initialized and optimized by themselves independently. Since object queries play the role of probing and pooling the features of certain instances, they have inconsistent conc entration preferences in different models. Consequently, the formulation of DETR naturally lacks the ability to provide sufficient distillation points with strict consistency between teacher and student models, and the predictions acquired from teacher are not informative or reliable for student to mimic.

\subsection{Distillation with Inconsistent Distillation Points}
To validate the analysis above that the sufficiency and consistency of distillation points is the essential challenge in DETR distillation, we start with applying the original logit-based distillation method\cite{chen2017learning} in classic detector to DETR, which mimics the category and box location logits predictions of teacher model. 

We examine three distillation points strategies: \textit{Inconsistent}, \textit{Similar Foreground}, and \textit{Similar General}. In \textit{Inconsistent}, all the object queries are viewed as distillation points with their original permutation; In \textit{Similar Foreground}, only object queries matched to ground truth in bipartite matching will be used as distillation points, and will permute with the same order of ground truth label; \textit{Similar General} further increase the number of distillation points by viewing the average of negative object queries in bipartite matching as a general background distillation point.  Experiments are conducted on DAB-DETR and evaluated on MS COCO2017, with ResNet18 as student and ResNet-50 as teacher. 

As shown in Table \ref{inconsistent}, \textit{Inconsistent} distillation points results in great degradation of the student model with unreliable knowledge from teacher model; \textit{Similar Foreground} with semantic-similar foreground distillation points alleviate the problem; and \textit{Similar General} achieves further improvement by increasing the number of distillation points with general background features. The preliminary experiments validate that the sufficiency and consistency of distillation points are of prime importance to improve the performance of student model in DETR distillation. 
\vspace{-5pt}
\section{KD-DETR}
\begin{figure*}[t]
  \centering
   \includegraphics[width=0.8\linewidth]{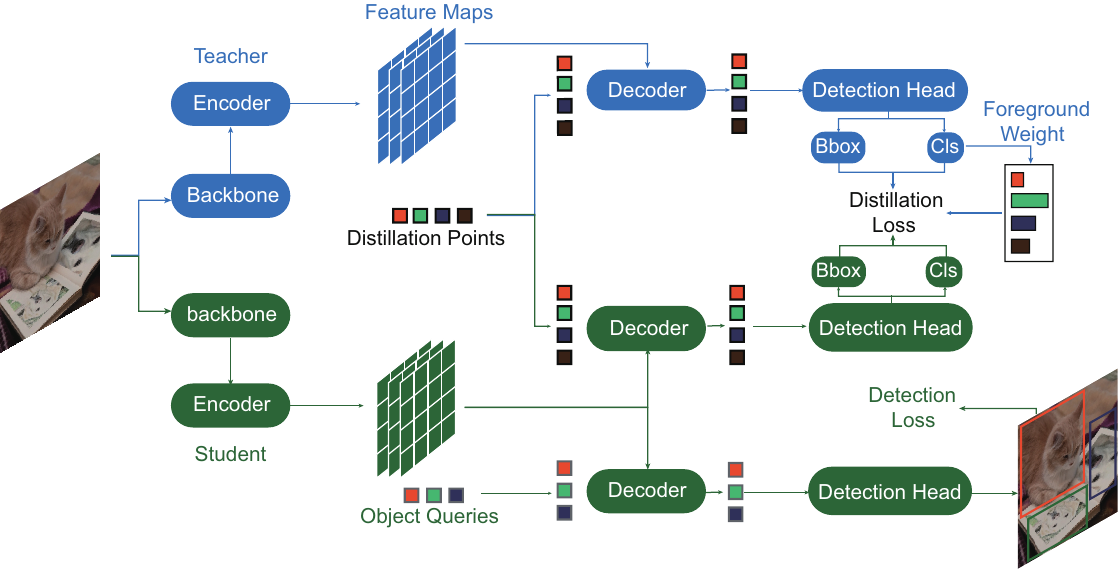}
   \caption{\textbf{KD-DETR architecture} KD-DETR decouples the distillation and detection tasks by introducing a set of distillation points shared between teacher and student models. For distillation, the distillation points are served as the query of the transformer decoders for both teacher and student models, and the student will mimic teacher's classification and box location predictions. The weight of each distillation point is measured by its foreground probability predicted by teacher. For detection task, the original object queries are processed by the student decoder for final prediction.}
   \vspace{-5pt}
   \label{framework}
\end{figure*}
To address the lack of consistent distillation points in DETR, we propose a general knowledge distillation paradigm for DETR (KD-DETR) with consistent distillation points sampling. As illustrated in Figure \ref{framework}, KD-DETR introduces a set of specialized object queries $\mathbf{\tilde{q}}$ shared between teacher and student models to construct distillation points. Decoupling the distillation task and detection task, KD-DETR can provide sufficient and consistent distillation points. We denote the original input and sampled distillation points as $\mathbf{x}=\{\mathbf{I}, \mathbf{q}\}, \mathbf{\tilde{x}}=\{\mathbf{I}, \mathbf{\tilde{q}}\}$ respectively.

For detection task, the student model is first optimized by its original detection loss: the original input $\mathbf{x}$ is fed into student model to make category and box location predictions, which will be assigned to the ground truth with bipartite matching and calculate detection loss $ \mathcal{L}_{det}$.

For distillation task, the sampled distillation points $\mathbf{\tilde{x}}$ will be fed into both student and teacher to make category and location predictions $\mathbf{c}, \mathbf{b}$:
\vspace{-5pt}
\begin{align}
    & \mathbf{c^s}, \mathbf{b^s} = f^s(\mathbf{I}, \mathbf{\tilde{q}}), \\
    & \mathbf{c^t}, \mathbf{b^t} = f^t(\mathbf{I}, \mathbf{\tilde{q}}), 
\end{align}
where $f^s, f^t$ refer to student and teacher model respectively. The distillation loss is calculated with following form:
\vspace{-5pt}
\begin{equation}
\begin{split}
    \mathcal{L}_{distill}  = &\sum_{i=1}^{M} [\lambda_{cls} \mathcal{L}_{KL}(\mathbf{\hat{c}}^t_i\Vert \mathbf{\hat{c}}^s_i) + \lambda_{L1}\mathcal{L}_{L1}(\mathbf{b}_i^s, \mathbf{b}_i^t), \\
    +& \lambda_{GIoU}\mathcal{L}_{GIoU}(\mathbf{b}_i^s, \mathbf{b}_i^t) ], 
\label{distill_loss}
\end{split}
\end{equation}
where M denotes the number of distillation points. For classification, we choose the KL-divergence $\mathcal{L}_{KL}$ as distillation loss with temperature $T$: $\mathbf{\hat{c}} = SoftMax(\frac{\mathbf{c}}{T})$. For box regression, $\mathcal{L}_{L1}$ and $\mathcal{L}_{GIoU}$ represent the L1 and GIoU loss for location distillation, which have the same formulation with detection loss. $\lambda_{cls}, \lambda_{L1}, \lambda_{GIoU}$ represent the coefficient of corresponding loss.

The total Loss is calculated with following form:
\begin{equation}
\begin{split}
    \mathcal{L}  = \mathcal{L}_{distill} + \mathcal{L}_{det}
\label{loss}
\end{split}
\end{equation}

\subsection{Distillation Points Sampling}
Generally, object queries are a set of abstract features responsible for certain objects by probing and pooling the context features from encoder. Existing works interpret object queries as anchors or reference points, revealing that each object query is sensitive to a particular region on the feature maps. Following this perspective, we provide a comprehensive general-to-specific sampling strategies for distillation points sampling: $\mathbf{\tilde{q}} = \{\mathbf{q_g}, \mathbf{q_s}\}$. 

\noindent\textbf{General Sampling with Random Initialized Queries.} 
In general sampling, we hope to probe teacher's general responses on different locations of the features by sparsely scanning the whole feature maps. Therefore, we randomly initialize a set of object queries, which are uniformly distributed on the features, to construct the general distillation points: $\mathbf{q_g}=\{\mathbf{q_i}\sim\mathcal{U}(0, 1)|i=1,...,M_g\}$, where $M_g$ denotes the number of general distillation points. To learn more general knowledge from teacher, we leave these distillation points unlearnable during training, and re-sample them every iteration.

\noindent\textbf{Specific Sampling with Teacher Queries.} While the general sampling provides a global retrieval of the features, we further propose a specific sampling strategy, focusing on those regions where teacher pays more attention. An intuitive way for specific sampling is to directly reuse the well-optimized object queries in teacher model:$\mathbf{q_s}=\mathbf{q_{teacher}}$. While teacher model is learned to concentrate more on these object queries, the predictions in these areas are more precise and informative.

\noindent\textbf{Foreground Rebalance Weight.}
\begin{table*}[th]
\begin{center}
\begin{tabular}{cc|c|c|ccccc|cc}
\toprule[2pt]
Models &  & Epochs & AP & $AP_{50}$ & $AP_{75}$ & $AP_{S}$ & $AP_{M}$& $AP_{L}$ & GFlops & Params \\
\hline
DAB-DETR& ResNet-50(T)& 50 & 42.1 &63.1 & 44.7&21.5 &45.7  &69.3 & 90 & 44M\\
& ResNet-18(S)& 50 & 36.2 &56.1 & 37.9& 16.9&39.0 & 53.5& 49 & 23M\\
& Ours& 50 & 41.4 & 61.4& 44.2&21.2 &44.7 &58.7 &  49 &23M\\
& Gains&  & \textbf{+5.2} & \textbf{+5.3}& \textbf{+6.3}&\textbf{+4.3} &\textbf{+5.7} &\textbf{+5.2} & &\\
\hline
DAB-DETR& ResNet-101(T)& 50 & 43.5 &63.9 & 44.6& 23.6& 47.3& 61.5 & 157 & 63M\\
& ResNet-50(S)& 50 & 42.1 &63.1 & 44.7&21.5 &45.7  &60.3 & 90 & 44M \\
& Ours& 50 & 45.7 &66.3 &49.4 &26.4 &49.8 &62.7 & 90 & 44M\\
& Gains&  & \textbf{+3.6} &\textbf{+3.2} &\textbf{+4.7} &\textbf{+4.9} &\textbf{+4.1} &\textbf{+2.4} & &\\
\hline
Deformable-DETR& ResNet-50(T)& 50 & 44.5 &63.6 &48.7 & 27.1& 47.6& 59.6& 171 & 40M  \\
& ResNet-18(S)& 50 & 40.1 & 58.1&43.7 & 22.4& 42.8&54.2 & 127 & 24M\\
& Ours& 50 & 43.7 & 62.1& 47.7&25.9 & 46.8& 57.6 & 127& 24M\\
& Gains & & \textbf{+3.6} & \textbf{+4.0} & \textbf{+4.0} & \textbf{+3.5} & \textbf{+4.0} & \textbf{+3.4}\\
\hline
Deformable-DETR& ResNet-101(T)& 50 & 48.0 & 66.7&52.6 &30.5 &52.3 &  62.3 & 238 &59M\\
& ResNet-50(S)& 50 & 44.5 &63.6 &48.7 & 27.1& 47.6& 59.6& 171& 40M  \\
& Ours& 50 & 48.3 & 66.7 & 52.9 & 30.8 & 52.1 & 62.5 &171 &40M\\
& Gains & & \textbf{+3.8} & \textbf{+3.1} & \textbf{+4.2} & \textbf{+3.7} & \textbf{+4.5} & \textbf{+2.9}\\
\hline
DINO& ResNet-50(T)& 36 & 50.9 & 69.0 & 55.3 & 34.6& 54.1& 64.6 & 245 & 47M\\
& ResNet-18(S)& 12 & 44.0 & 61.2 & 48.1& 27.4 & 46.9 & 56.9& 200 & 30M\\
& Ours& 12 & 48.4 & 65.5 & 53.0 & 31.6 & 51.7 & 62.3 & 200 & 30M \\
& Gains&  & \textbf{+4.4} & \textbf{+4.3} & \textbf{+4.9} & \textbf{+4.2} & \textbf{+4.8} &\textbf{+5.4}& &\\
\hline
DINO &ResNet-101(T) & 36 & 51.3 & 69.5 & 55.8 & 34.8 &  54.8 & 65.8 & 311 & 66M\\
& ResNet-50(S)& 12 & 49.0 & 66.6 & 53.5 & 32.0 & 52.3 & 63.0 & 245 & 47M\\
& Ours& 12 & 51.6 & 69.6 & 56.6 & 34.2& 54.8 & 66.9 & 245 &47M\\
& Gains&  & \textbf{+2.6} & \textbf{+3.0} & \textbf{+3.1} & \textbf{+2.2} & \textbf{+2.5} &\textbf{+3.9}& &\\
\bottomrule[2pt]
\end{tabular}
\end{center}
\vspace{-5pt}
\caption{Results of the proposed KD-DETR with different DETR detectors and backbones with various scale.}
\vspace{-5pt}
\label{results}
\end{table*}
The imbalance between foreground and background regions is one critical problem in object detection distillation, not special in DETR. An intuitive way is to utilize the classification scores of distillation points which are predicted by teacher model to re-balance the distillation loss. Concretely, those distillation points with higher classification scores are regarded as foreground distillation points, containing more useful information for detection, and should be given more attention.
\begin{equation}
    w_i = \max_{c\in[0, K]}p^t(y_c|\mathbf{q}_i),
\end{equation}

\noindent where $p^t(y_c|\mathbf{q}_i)$ denotes the probability of $\mathbf{q}_i$ assigned to category $c$ predicted by teacher model, and $w_i$ denotes the foreground rebalance weight of $\mathbf{q}_i$. In this way, the distillation loss in Eq.~\eqref{distill_loss} will be writen as follow:
\begin{equation}
\begin{split}
    \mathcal{L}_{distill}  = &\sum_{i=1}^{M} w_i[\lambda_{cls} \mathcal{L}_{KL}(\mathbf{\hat{c}}^t_i\Vert \mathbf{\hat{c}}^s_i) \\
    + &\lambda_{L1}\mathcal{L}_{L1}(\mathbf{b}_i^s, \mathbf{b}_i^t) 
    + \lambda_{GIoU}\mathcal{L}_{GIoU}(\mathbf{b}_i^s, \mathbf{b}_i^t) ] 
\label{distill_coarse}
\end{split}
\end{equation}

\subsection{Generalization to Heterogeneous Distillation}
To further extend the generalization of the KD-DETR paradigm, we apply the idea to heterogeneous distillation between DETR and CNN detector. Intuitively, both the anchor in the CNN detector and the object query in DETR represent certain locations on the image and share spatial consistency. We first construct a set of distillation points $ \mathbf{\tilde{q}}$ with the coordinate of anchors in CNN detector, and then convert them to the formulation of object queries in DETR. The distillation loss is in the formulation of Eq.~\eqref{distill_loss}.

We further propose a simple but effective strategy for distillation points sampling in heterogeneous distillation with Intersection over Union (IoU). Specifically, we calculate the IoU between anchors and ground truth grounding boxes, and select the top $k$ anchors as distillation points. Details about heterogeneous distillation are in Appendix \ref{sec:het}.

\section{Experiment}

To validate the effectiveness and generalization of KD-DETR, we evaluate it on different DETR architectures including DAB-DETR, Deformable DETR and DINO, with two scales of backbones: ResNet-50 and ResNet18. We also extend KD-DETR to heterogeneous distillation, and evaluate on distillation between DINO-Res50 and Faster-RCNN Res50. To support our analysis of consistent distillation points, we further conduct an extensive ablation study. 

\subsection{Experimental Settings}

\noindent\textbf{Datasets}: All the proposed experiments are evaluated on MS COCO2017\cite{lin2014microsoft} spanning 80 categories, with the default split of ~117k training images for training and 5k validate images for testing.  Standard COCO evaluation metrics are adopted for validation.

\noindent\textbf{Implementation Details}: As KD-DETR is a plug-and-play distillation module, we follow the original settings of hyper-parameters and optimizer of all the student model for the training of detection part. We choose ResNet\cite{he2016deep} as backbone, which are pre-trained on ImageNet1K\cite{krizhevsky2017imagenet}. We propose the inheriting strategy\cite{kang2021instance} to initialize students' level embeddings on multi-scale DETR (details in Appendix \ref{sec:inherit}). For distillation task, we set hyper-parameters of the coefficient of the distillation loss as $\lambda_{kl}=1, \lambda_{L1}=5, \lambda_{GIoU}=2$. The number of General distillation points is $300, 300, 900$ for DAB-DETR, Deformable DETR, and DINO. For heterogeneous distillation, the hyper-parameter $k$ for IoU sampling is set to $10$.  We train our models on Nvidia A100 GPUs with batch size set to 16 in total. 

\begin{table}[t]
\begin{center}
\begin{tabular}{c|ccccc}
\toprule[2pt]
Method & AP & $AP_{S}$ & $AP_{M}$ & $AP_{L}$   \\
\hline
Deformable DETR Res50 & 44.5 & 27.1 & 47.6	 & 59.6  \\
FGD\cite{yang2022focal} & 44.1 & 25.9 &  47.7 & 58.8  \\
FitNet\cite{romero2014fitnets} & 44.9 & 27.2 & 48.4 & 59.6 \\
DETRDistill\cite{chang2023detrdistill} & 46.6 & 28.5 & 50.0 & 60.4 \\
\hline
Ours & \textbf{48.3}&  \textbf{30.8} &  \textbf{52.1}&  \textbf{62.5}  \\
\bottomrule[2pt]
\end{tabular}
\end{center}
\caption{Comparison with state-of-the-art on Deformable DETR.}
\vspace{-5pt}
\label{sota}
\end{table}

\subsection{Distillation on Different DETR benchmarks}
We evaluate our method on three typical DETR architectures: DAB-DETR\cite{liu2021dab}, a single-scale DETR with special object query settings; Deformable-DETR\cite{zhu2020deformable}, a multi-scale DETR with deformable attention module; and DINO\cite{zhang2022dino}, which combines a series of novel techniques including deformable attention, two-stage object queries settings, and a de-noising module, to evaluate our effectiveness on well-optimized model with high accuracy. Distillation results of DINO with Swin Transformer backbone and compressing the layers of DETR are in Appendix \ref{sec:backbone}, \ref{sec:layers}

The results are illustrated in Table \ref{results}. For DAB-DETR, KD-DETR significantly boosts the performance of ResNet-18, and ResNet-50 with $5.2\%$, and $3.6\%$ mAP improvement respectively. Note that student with ResNet-50 even surpass the teacher by $2.2\%$ mAP. For Deformable DETR with ResNet-18 and ResNet-50, KD-DETR achieve $3.6\%$ and $3.8\%$ mAP gains, and student with ResNet-50 also exceed the teacher with $0.3\%$ mAP. For DINO with ResNet-18 and ResNet-50, KD-DETR also improves $4.4\%$ and $2.6\%$ mAP of the student models on 12-epoch training scheduler. 

Table \ref{sota} shows the comparison with state-of-the-art distillation method. With simple logit-based distillation alone, KD-DETR outperforms former methods , with an improvement of $1.7\%$ on Deformable DETR with ResNet-50 compared with DETRDisitll\cite{chang2023detrdistill}



\subsection{Generalization to Heterogeneous Distillation}

Table \ref{cad} presents the results with Faster RCNN ResNet-50 as student and DINO ResNet-50 as teacher. KD-DETR works well on heterogeneous distillation task, improving the student models with $2.1\%$ mAP, achieving the performance of state-of-the-art homogeneous methods. This validate our analysis on the effect of distillation points on knowledge distillation, and bridge the gap of transferring knowledge between detectors with different architectures. 
\begin{table}[t]
\begin{center}
\begin{tabular}{c|cccc}
\toprule[2pt]
Method & AP & $AP_{S}$ & $AP_{M}$ & $AP_{L}$   \\
\hline
RCNN-Res50 & 38.4 & 21.5 & 42.1 & 50.3  \\
FGFI\cite{wang2019distilling}& 39.3 & 22.5 & 42.3 & 52.2  \\
GID\cite{dai2021general} &  40.2&22.7 & 44.0& 53.2 \\
FGD\cite{yang2022focal} & 40.4 & \textbf{22.8}&  44.5 & 53.5  \\
\hline
Ours & \textbf{40.5} & 22.7 & \textbf{44.6}& \textbf{53.8}   \\
\bottomrule[2pt]
\end{tabular}
\end{center}
\caption{\textbf{Heterogeneous Distillation}: Distillation between DINO and Faster RCNN}
\label{cad}
\end{table}

\subsection{Ablation Study}
The ablation study is conducted on MS COCO2017 with DAB-DETR with backbone of ResNet50 as teacher and ResNet18 as student. 

\noindent\textbf{Analysis on the Benefit of Knowledge Distillation}

In KD-DETR, we introduce a set of specialized distillation points in the training process to the original DETR architecture. \cite{zhang2022dino}\cite{li2022dn} have proven that the number of object queries will affect the performance of model. To validate that the benefit is from knowledge distillation rather than the increase of object queries, we conduct an ablation experiment by applying an additional set of object queries with the same number of distillation points.

As shown in Table \ref{benefit}, simply increasing the number of object queries only achieves trivial gains, and the main contribution of boosting the model's performance is from the knowledge transferred from teacher with distillation points.

\begin{table}[t]
\begin{center}
\begin{tabular}{cc|ccc}
\toprule[2pt]
\makecell[c]{Object\\ Queries} & \makecell[c]{Distillation \\ Points}& AP & $AP_{50}$ & $AP_{75}$  \\
\hline
300 & -  & 36.2   & 56.1 &  37.9 \\
300+300 & - & 37.7  &58.0 &  41.1 \\
300 & 300 & 40.2  & 60.7 & 42.8 \\
\bottomrule[2pt]
\end{tabular}
\end{center}
\vspace{-5pt}
\caption{\textbf{Benefit of Knowledge Distillation}: Simply increasing the set of object queries leads to trivial improvement}
\vspace{-5pt}
\label{benefit}
\end{table}

\noindent\textbf{Analysis on the Distillation Points Sampling Strategy}

A comprehensive general-to-specific distillation points sampling scheme is introduced in this paper, including three strategies: General Sampling, Specific Sampling, and Foreground Rebalance Weight. 
The ablation results are illustrated in Table \ref{strategy}. The general sampling strategy with randomly initialized queries can boost the performance of student model for $2.5\%$ mAP, while the specific sampling strategy with teacher queries achieves an improvement of $3.8\%$. When refining the general distillation points with foreground rebalance weight, the general sampling strategy and specific sampling strategy yield larger gains for $3.7\%$ and $4.0\%$, respectively. The combination of the three strategies further promotes the performance for $5.2\%$.

It is also important to note that the performance of general sampling with foreground rebalance weight is fairly close to the specific sampling. Since the teacher queries in specific sampling are well-optimized and concentrate more on the foreground regions, such phenomena validate that the general sampling with randomly initialized queries can probe the whole feature maps evenly, and foreground rebalance weight can help the student focus more on the foreground regions. In addition, the further improvement brought by the combination of three strategies indicates that the specific distillation points bring more information from the concentration of teacher model.
\begin{table}[t]
\begin{center}
\begin{tabular}{ccc|ccc}
\toprule[2pt]
General & Specific & FRW & AP & $AP_{50}$ & $AP_{75}$  \\
\hline
&&& 36.2 & 56.1 & 37.9 \\
$\checkmark$&  &  & 38.7 & 59.1 & 40.7 \\
$\checkmark$& & $\checkmark$ & 39.9 & 60.2& 42.5  \\
& $\checkmark$ &  & 40.0 & 60.5 & 42.4  \\

& $\checkmark$& $\checkmark$ & 40.2 &60.7 & 42.8 \\
$\checkmark$& $\checkmark$& $\checkmark$ & 41.4 &61.4 &44.2 \\
\bottomrule[2pt]
\end{tabular}
\end{center}
\vspace{-5pt}
\caption{Distillation with Different Distillation Points Sampling Strategies. FRW refers to Foreground Rebalance Weight.}
\vspace{-5pt}
\label{strategy}
\end{table}

\noindent\textbf{Number of General Distillation Points}

We investigate the influence of the number of general distillation points by varying the number from 10 to 900. As is shown in Table \ref{number}, the improvement is significant when increasing the distillation points from 10 to 50, while gradually saturating when continuing to increase to 300. There is even a slight degradation when further increasing the number. These phenomena validate that the general distillation points can effectively probe the knowledge from teacher's attention in different regions when sparsely distributed on the feature maps. However, dense distillation points will introduce more noise of background information, and will be harmful for distillation.
\begin{table}[t]
\begin{center}
\begin{tabular}{c|ccccc}
\toprule[2pt]
Number & 10 & 50 & 100 & 300 & 900  \\
\hline
AP             & 38.6 & 39.3  & 39.5  & 39.9 & 39.5\\
$AP_{50}$& 58.3 & 59.2  & 59.3 & 60.2& 60.4\\
$AP_{75}$& 41.0 & 41.3  & 41.8 &  42.5 & 41.7 \\
\bottomrule[2pt]
\end{tabular}
\end{center}
\vspace{-5pt}
\caption{Distillation with Different Number of General Distillation Points}
\vspace{-10pt}
\label{number}
\end{table}

\noindent\textbf{Visualization of the Distillation Points}

\begin{figure}[t]
  \centering
   \includegraphics[width=1\linewidth]{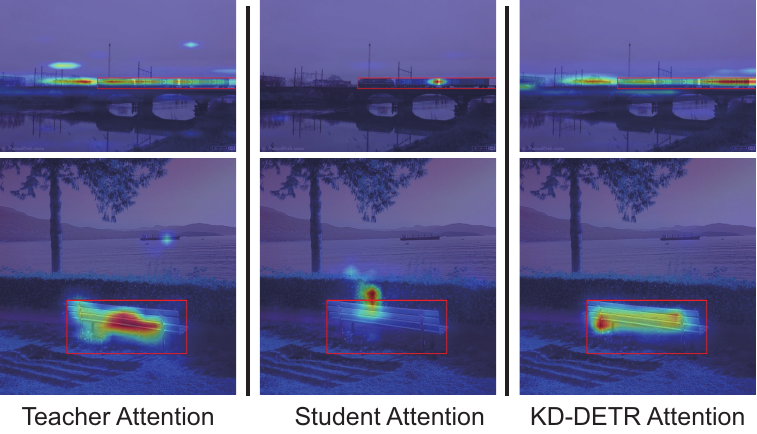}
  \caption{Attention Map Visualization of Specific Distillation Points: Images from left to right are from teacher, original student and student with KD-DETR. The corresponding predicted bounding box of the distillation points are marked with red rectangles. }
   \vspace{-5pt}
   \label{specific}
\end{figure}

\begin{figure}[t]
  \centering
    
    \begin{minipage}{0.32\linewidth}
  \centering
   \includegraphics[width=1\textwidth]{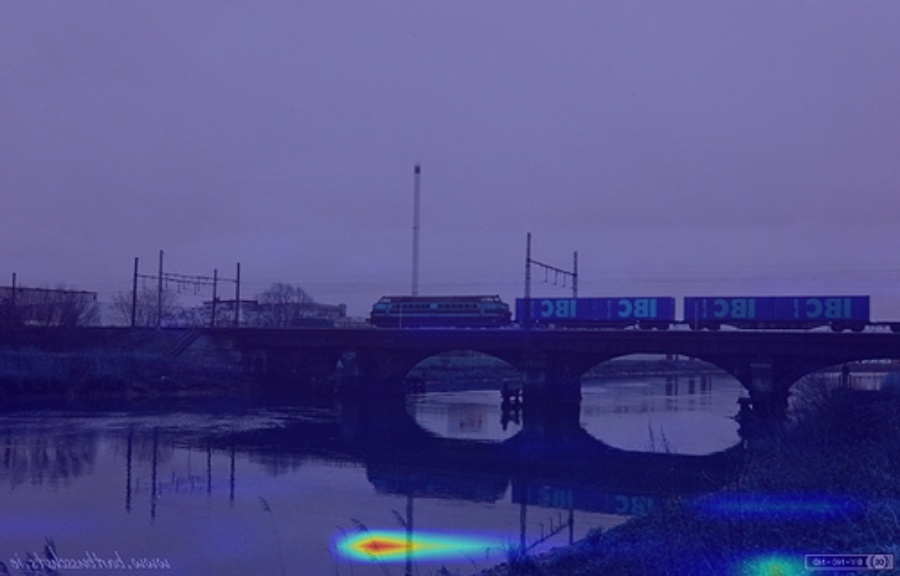}
    \label{fig:short-a}
  \end{minipage}
  \begin{minipage}{0.32\linewidth}
  
  \centering
   \includegraphics[width=1\textwidth]{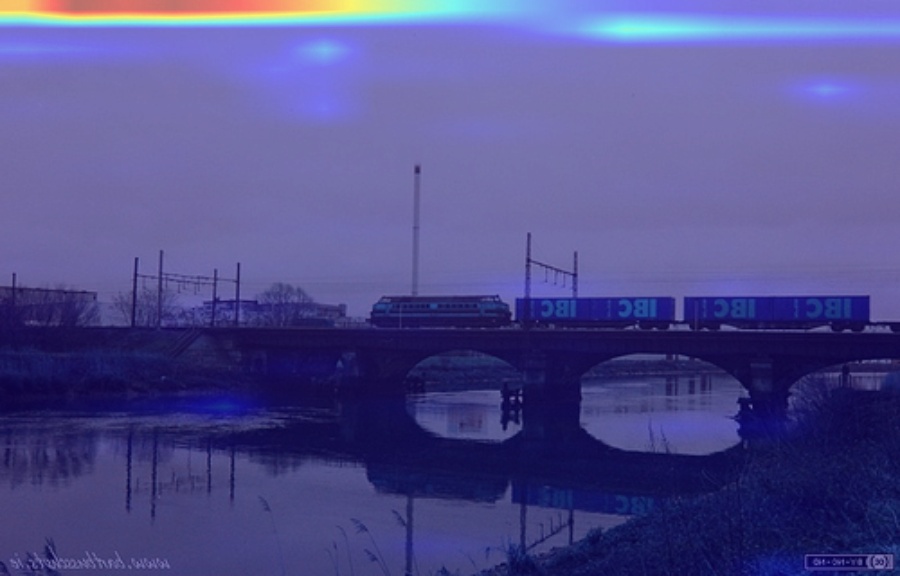}  
    \label{fig:short-b}
  \end{minipage}
   \begin{minipage}{0.32\linewidth}
   \centering
   \includegraphics[width=1\textwidth]{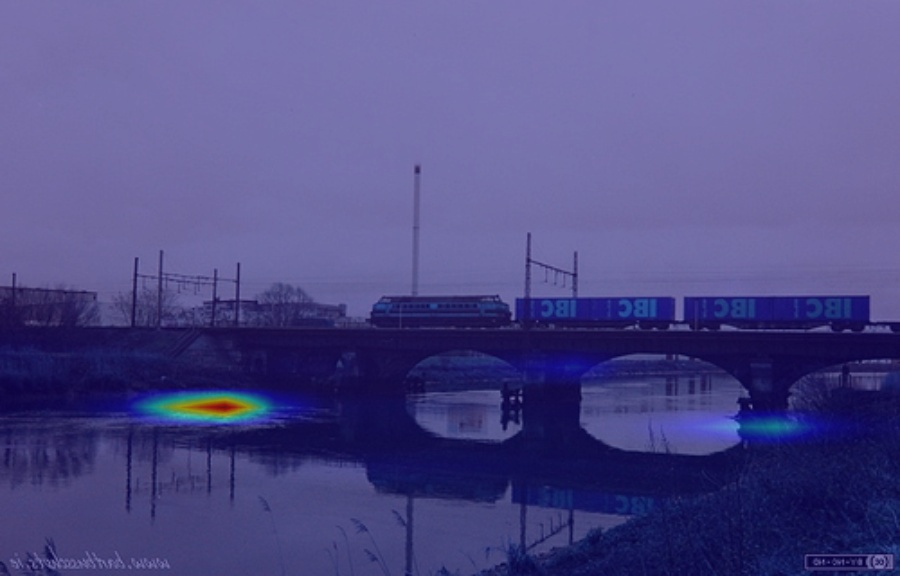}  
    \label{fig:short-b}
  \end{minipage}\\ 
       \begin{minipage}{0.32\linewidth}
  \centering
  \vspace{-10pt}
   \includegraphics[width=1\textwidth]{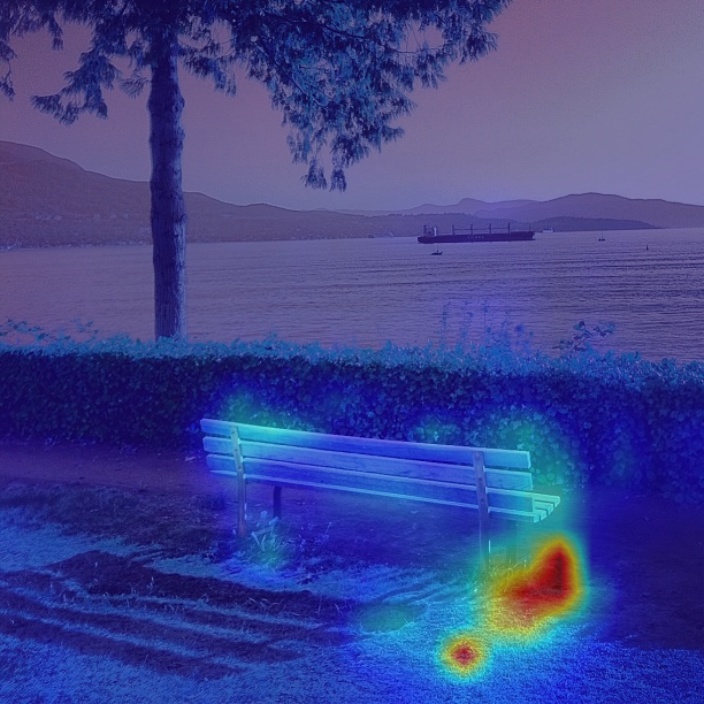}
    \label{fig:short-a}
  \end{minipage}
  \begin{minipage}{0.32\linewidth}
  \centering
  \vspace{-10pt}
   \includegraphics[width=1\textwidth]{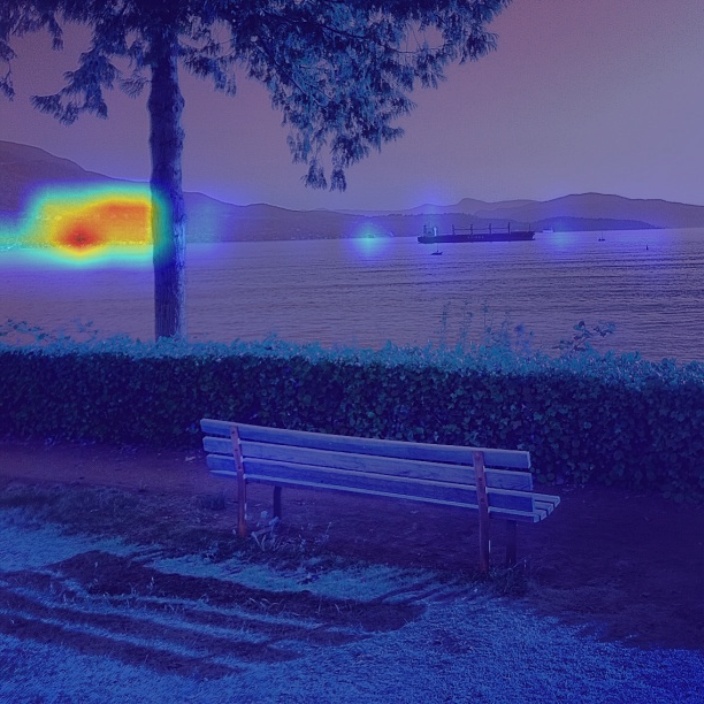}  
    \label{fig:short-b}
  \end{minipage}
   \begin{minipage}{0.32\linewidth}
   \centering
   \vspace{-10pt}
   \includegraphics[width=1\textwidth]{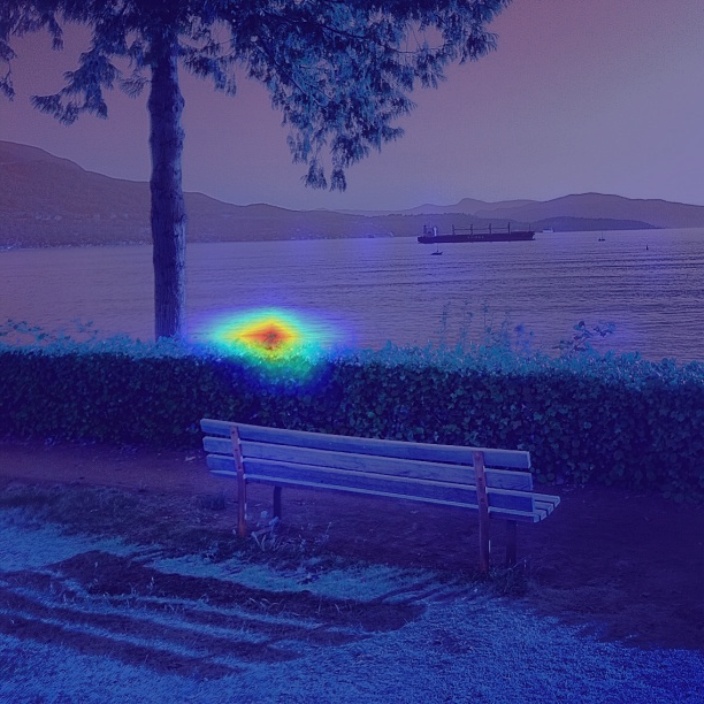}  
    \label{fig:short-b}
  \end{minipage}\\ 
  \vspace{-10pt}
 \caption{Attention Map Visualization of General Distillation Points: Teacher model's attention of general distillation points focuses more on the background regions, providing more information for student to mimic.}
 \vspace{-10pt}
  \label{general}
\end{figure}

In order to understand what knowledge has been transferred from teacher to student better, we visualize the attention map of different kinds of distillation points. Figure \ref{specific} illustrates the attention of the same specific distillation points from teacher model, student model and KD-DETR. It can be seen that the specific distillation points focus more on the features of foreground regions. While the attention of the original student model without distillation is different from the teacher model, KD-DETR can align the concentration of teacher and student models. On the contrary, the general distillation points, as illustrated in Figure \ref{general}, concentrate more on the background regions, and objects that are not included in the ground truth annotations, which can provide more additional semantic information to the student. In this way, the combination of general and specific distillation points can provide more comprehensive knowledge.

\section{Conclusion}

In this paper, we study the compression of DETR with knowledge distillation. We provide thorough experimental and theoretical analysis on the key point in DETR distillation. Based on the analysis, we propose the first general knowledge distillation paradigm for DETR (KD-DETR), together with a comprehensive general-to-specific consistent distillation sampling scheme. We conduct extensive experiments to demonstrate the flexibility, generalization, and extensibility of KD-DETR on various DETR architectures, and for both homogeneous and heterogeneous distillation. For homogeneous distillation, KD-DETR compresses both the scale of backbone and transformer layers, significantly boosting the performance of student model. For heterogeneous distillation, KD-DETR effectively transfers the knowledge from DETR to CNN detector.

{
    \small
    \bibliographystyle{ieeenat_fullname}
    \bibliography{main}
}
\clearpage
\setcounter{page}{1}
\maketitlesupplementary
\section{Heterogeneous Distillation} 
\label{sec:het}
\begin{figure}[t]
  \centering
   \includegraphics[width=1.0\linewidth]{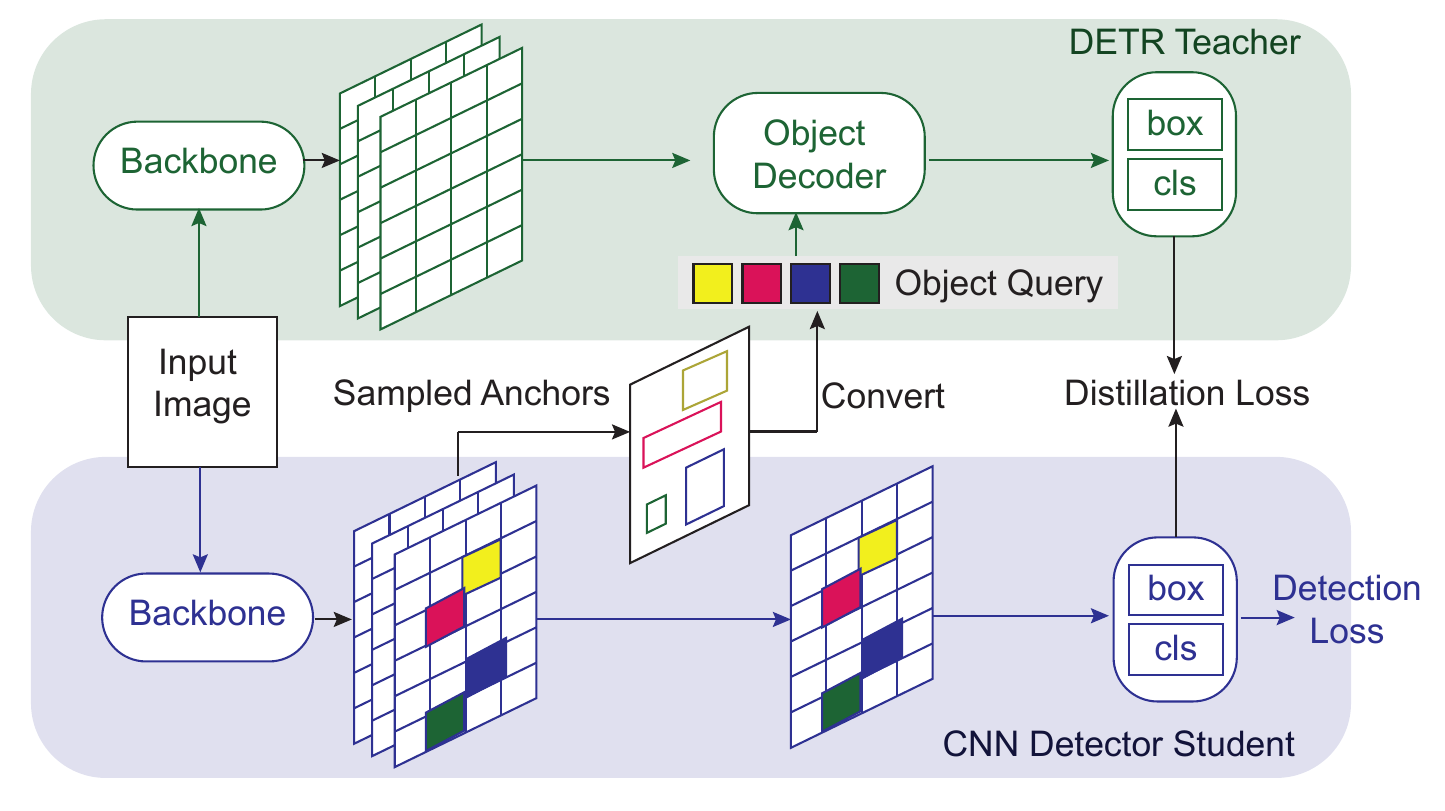}
   \caption{\textbf{Heterogeneous Distillation}}
   \label{het_framework}
\end{figure}
In heterogeneous distillation, the crucial part is to construct consistent distillation points between DETR and CNN-detector, indicating object queries and anchors respectively. KD-DETR propose the first idea in heterogeneous distillation by constructing the consistency between the object query and the anchor via spacial coordination: the anchor is generated through the sliding window strategy, and can be represented as $A=\{x_a, y_a, w_a, h_a\}$; while the object queries in most DETR, including DINO, are generated from anchor boxes: $Q=MLP(PE(cx, cy, w, h))$, where $PE$ is positional encoding and $MLP$ refers to a MLP projector. In this way, the anchor can be directly converted into the object query, and utilized as consistent distillation points:
\begin{small}
\begin{equation}
Q_A=MLP(PE(x_a + \frac{w_a}{2}, y_a+\frac{h_a}{2}, w_a, h_a))
\label{eq}
\end{equation}
\end{small}

\noindent As shown on Figure \ref{het_framework}, KD-DETR constructs distillation points by sampling anchors generated in CNN-detector (sampling details in Sec.4.2), then convert them to object queries of DETR via Eq.~\ref{eq}.  With the predictions of distillation points from student and teacher, the distillation loss is Eq.3 and the total loss is  Eq.4.

\section{More Ablation Study and Analysis}
\subsection{Inheriting Stratgy}
\label{sec:inherit}
For DETR with multi-scale features, including Deformable DETR and DINO, we propose the inheriting strategy\cite{kang2021instance} by initialize the student's level embeddings with teacher's parameters. As shown in Table \ref{inherit}, inheriting strategy brings an additional $0.3\%$ promotion on Deformable DETR Res18. Such phenomena also validate our analysis on consistent distillation points, as the level embeddings in DETR is a set of learnable embeddings for model to distinguish different scale of features, and are egocentirc. That is to say, on multi-scale DETR, the formulation of distillation points turns to $x=(I+LE, q)$, where $LE$ denotes the level embeddings. In this way, inheriting the level embeddings from teacher to student can restrict the consistency of distillation points.
\begin{table}[t]
\begin{center}
\begin{tabular}{cc|cccc}
\toprule[2pt]
Model& Arch & AP & $AP_{50}$ & $AP_{75}$    \\
\hline
Deformable DETR & Res50 & 44.5 & 63.6 & 52.6  \\
Deformable DETR & Res18 & 40.1 & 58.1& 43.7  \\
Ours & Res18 & 43.4 & 61.8 & 47.5  \\
Ours$\dagger$ & Res18 & \textbf{43.7} & \textbf{62.1} & \textbf{47.7}  \\
\bottomrule[2pt]
\end{tabular}
\end{center}
\caption{Level Embedding Inheriting Strategy on Deformable DETR. $\dagger$ means using inheriting strategy}
\label{inherit}
\end{table}
\subsection{Generalization on Advanced Backbone}
\label{sec:backbone}
To validate the extensibility of KD-DETR, we conduct additional experiments with DINO Swin Transformer[] as backbone. As shown in Table \ref {swin}, with a strong baseline, KD-DETR significantly boosts the performance of student models. For Swin-Tiny as student and Swin-Base as teacher, KD-DETR promotes the student’s COCO mAP from $50.7\%$ to $52.6\%(+1.9\%)$; For Swin-Base as student and Swin-Large as teacher, KD-DETR promotes student from $55.6\%$ to $57.1\% (+1.5\%)$.
\begin{table}[t]
\begin{center}
\begin{tabular}{cc|ccc}
\toprule[2pt]
Student& Arch & AP & $AP_{50}$ & $AP_{75}$    \\
\hline
DINO & Swin-T & 50.7 & 67.9 & 55.0 \\
Ours & Swin-T &52.6 & 70.3& 57.5  \\
Gains&  & \textbf{+1.9} & \textbf{+2.4} & \textbf{+2.5}  \\
DINO & Swin-B & 55.6 & 74.3 & 60.8 \\
Ours & Swin-B & 57.1 & 75.5 & 62.5  \\
Gains&  & \textbf{+1.5} & \textbf{+1.2} & \textbf{+1.7}  \\
\bottomrule[2pt]
\end{tabular}
\end{center}
\caption{Distillation on DINO with Swin Transformer backbone}
\label{swin}
\end{table}

\label{sec:inherit}
\subsection{Distillation on the Transformer Layers}
\label{sec:layers}
Besides the scale of backbone, the layer number of transformer encoder and decoder is also an important factor of the model size and computation cost in DETR. In this paper, we also conduct experiments to compress the layer numbers of transformer to validate the scalability of KD-DETR. The FPS reported is measured on a single Nvidia A100 GPU.

Table \ref{layer} shows the results of KD-DETR on DAB-DETR, with backbone of ResNet-50 as teacher and ResNet-18 as student. While decreasing the number of transformer layers will cause great degradation in the performance, KD-DETR can significantly boost the student model. For example, the student model with 2 encoder layers and 6 decoder layers can outperform the full-scale model for $2.8\%$ mAP with 1.2x FPS improvement.
\begin{table}[t]
\begin{center}
\begin{tabular}{c|ccc|cc}
\toprule[2pt]
Enc/Dec & AP & $AP_{50}$ & $AP_{75}$ & FPS & Params  \\
\hline
6/6 & 36.2 & 56.1 & 37.9 & 76 & 31M \\
Ours & 41.4(\textbf{+5.2}) & 61.4 & 44.2 & 76 & 31M \\
2/6 &  36.2&56.3 & 38.4 & 102 & 27M\\
Ours & 39.0(\textbf{+2.8}) & 58.9&  41.7 & 102 & 27M \\
6/2 &31.8 & 49.8&  33.5 & 82 & 25M\\
Ours & 38.9(\textbf{+7.1})& 58.0& 41.6 & 82 & 25M \\
2/2 & 29.3 & 46.6 & 37.0 & 113 &17M \\
Ours & 32.8(\textbf{+3.5}) &  52.6 & 34.2 &113 &17M\\
\bottomrule[2pt]
\end{tabular}
\end{center}
\caption{\textbf{Distillation on Transformer Layers}: Compressing the number of encoder layers and decoder layers with KD-DETR}
\label{layer}
\end{table}


\end{document}